\documentclass[conference]{IEEEtran}
\IEEEoverridecommandlockouts
\usepackage{cite}
\usepackage{blindtext}
\usepackage{amsmath,amssymb,amsfonts}
\usepackage{algorithm}%
\usepackage{algorithmicx}%
\usepackage{graphicx}
\usepackage{textcomp}
\usepackage{float}  
\usepackage{xcolor}
\usepackage{multirow}%
\usepackage{amsmath,amssymb,amsfonts}%
\usepackage{amsthm}%
\usepackage{mathrsfs}%
\usepackage{xcolor}%
\usepackage{textcomp}%
\usepackage{manyfoot}%
\usepackage{booktabs}%
\usepackage{algpseudocode}%
\usepackage{listings}%
\usepackage{amsthm} 
\usepackage{setspace}
\newtheorem{assumption}{Assumption}
\newtheorem{principle}{Principle}
\usepackage[colorlinks=true,
            linkcolor=magenta, 
            anchorcolor=black, 
            citecolor=black, 
            urlcolor=blue
            ]{hyperref}

\makeatletter

\begin{document}
\singlespacing 
\title{Grasping by Spiraling: Reproducing Elephant Movements with Rigid-Soft Robot Synergy}


\author{Huishi Huang$^{1,4*\dagger}$, Haozhe Wang$^{1,2*}$, Chongyu Fang$^{1\dagger}$, Mingge Yan $^{1}$, Ruochen Xu$^{1}$, Yiyuan Zhang$^{1}$,\\ Zhanchi Wang$^{3}$, Fengkang Ying$^{1,2}$ Jun Liu$^{4}$, Cecilia Laschi$^{1}$, and Marcelo H. Ang Jr.$^{1}$
\thanks{$^*$The authors contributed equally to this work.}
\thanks{$^\dagger$Corresponding Author.}%
\thanks{$^{1}$Authors are with the Advanced Robotics Centre, National University of Singapore, 117608, Singapore. {\tt\small \{huishi.huang, wang\_haozhe, fangchongyu\_14, mingge\_yan, xu\_ruochen, yiyuan.zhang, fengkang\}@u.nus.edu}, {\tt\small \{mpeclc, mpeangh\}@nus.edu.sg}}%
\thanks{$^{2}$Haozhe Wang and Fengkang Ying are with the Integrative Sciences and Engineering Programme, National University of Singapore Graduate School, 119077, Singapore.}%
\thanks{$^3$ Zhanchi Wang is with the School of Computer Science and Technology, University of Science and Technology of China, Hefei, Anhui 230027, China {\tt\small zkdwzc@mail.ustc.edu.cn}}%
\thanks{$^4$Huishi Huang and Jun Liu are with the Institute of High Performance Computing (IHPC), Agency for Science, Technology and Research (A*STAR), 138632, Singapore {\tt\small liuj@ihpc.a-star.edu.sg}}
}


\maketitle

\begin{abstract}
The logarithmic spiral is observed as a common pattern in several living beings across kingdoms and species. Some examples include fern shoots, prehensile tails, and soft limbs like octopus arms and elephant trunks. In the latter cases, spiraling is also used for grasping. Motivated by how this strategy simplifies behavior into kinematic primitives and combines them to develop smart grasping movements, this work focuses on the elephant trunk, which is more deeply investigated in the literature. We present a soft arm combined with a rigid robotic system to replicate elephant grasping capabilities based on the combination of a soft trunk with a solid body. In our system, the rigid arm ensures positioning and orientation, mimicking the role of the elephant's head, while the soft manipulator reproduces trunk motion primitives of bending and twisting under proper actuation patterns.



This synergy replicates 9 distinct elephant grasping strategies reported in the literature, accommodating objects of varying shapes and sizes. The synergistic interaction between the rigid and soft components of the system minimizes the control complexity while maintaining a high degree of adaptability. 


\end{abstract}


\section{Introduction}
\label{sec:intro}
In nature, logarithmic spirals are commonly observed in biological structures, combining remarkable aesthetic appeal with functional efficiency. A logarithmic spiral, also known as the equiangular spiral, is a self-similar curve in which the angle between the tangent to the spiral and the radial line from the center is constant. The polar equation for a logarithmic spiral is:

\begin{equation}
r = ae^{b\theta}
\end{equation}

where $r$ is the radius, $\theta$ is the polar angle, and $a$ and $b$ are real constants. The spiral expands logarithmically with $\theta$, resulting in an ever-widening curve without bounds. This pattern is a recurring motif across various prehensile appendages, such as the tails of chameleons and seahorses, the trunks of elephants, and even the human hand and arm \cite{gupta1998motion, littler1973adaptability}. These structures not only exhibit logarithmic spiral geometries in their static configurations but also leverage the dynamic properties of these spirals during grasping and manipulation tasks.

Static logarithmic spirals are evident in the resting morphology of many prehensile appendages. For example, animals often store their appendages in spiral configurations, tightly wrapping each segment to optimize space usage. This compact form could be an efficient way to conserve energy and maintain readiness for rapid deployment. In contrast, dynamic spirals emerge during the prehension process, where the inward propagation of curvature plays a key role in anchoring objects and transporting them securely. This behavior, particularly observed in the elephant trunk \cite{dagenais2021elephants}, demonstrates a proximal propagation of curvature during grasping and its reversal during object release. Such a motion strategy ensures efficient object handling, facilitating secure capture, transportation, and precise placement.

Elephant trunks exhibit remarkable versatility while maintaining a high force capacity. This is enabled by the activation of approximately 90,000 muscle fascicles \cite{longren2023dense}, which work in coordination to produce complex behaviors. They are arranged in a muscular hydrostat, comprising of longitudinal, transverse and oblique muscles, which keeps constant volume during contractions \cite{kier2012diversity}. In principle, a muscular hydrostat like the trunk could exhibit a virtually infinite number of degrees of freedom. However, to simplify control for specific tasks, elephants have evolved strategies that involve identifying the most efficient kinematic primitives and combining them in finite configurations to address diverse scenarios \cite{dagenais2021elephants}.

\begin{figure*}[htbp]
\centering
\includegraphics[page=1, width=\textwidth]{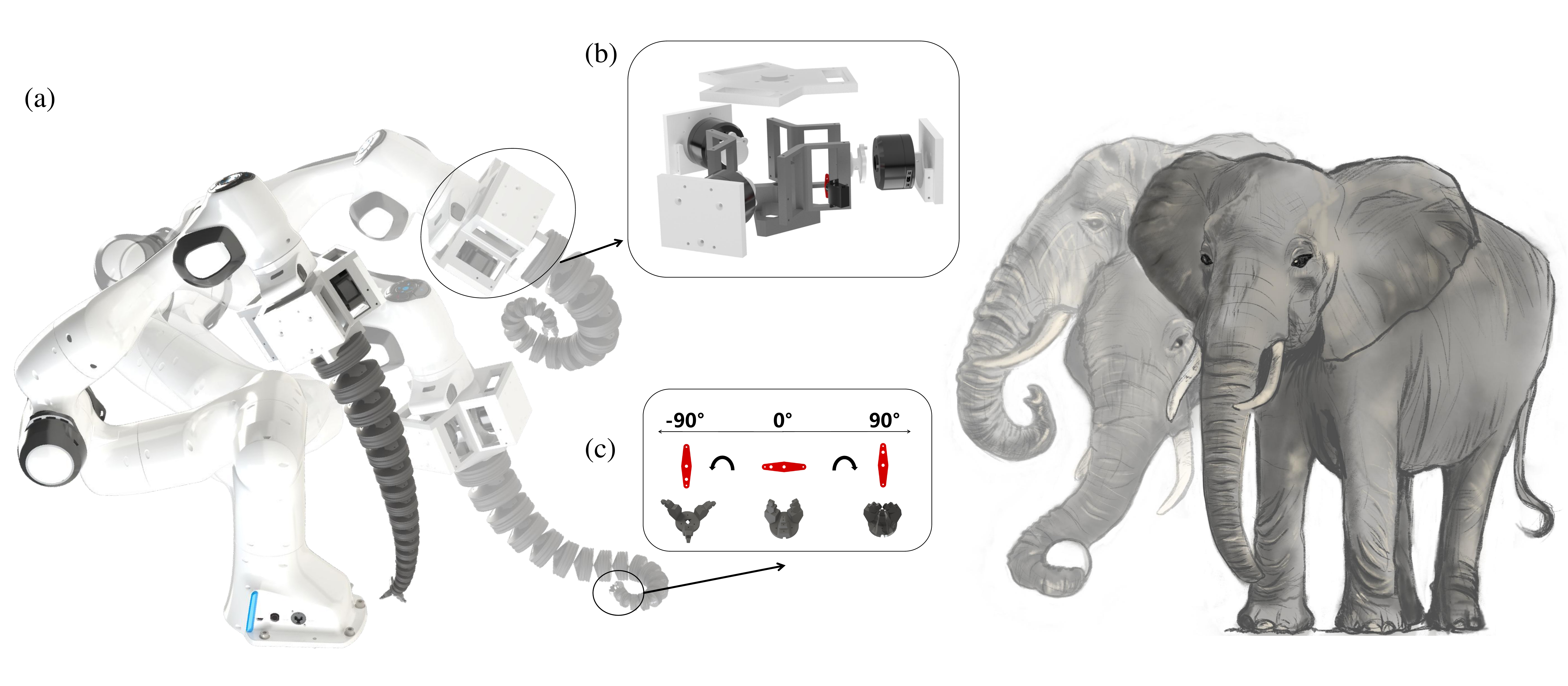}\\
\includegraphics[page=1, width=\textwidth]{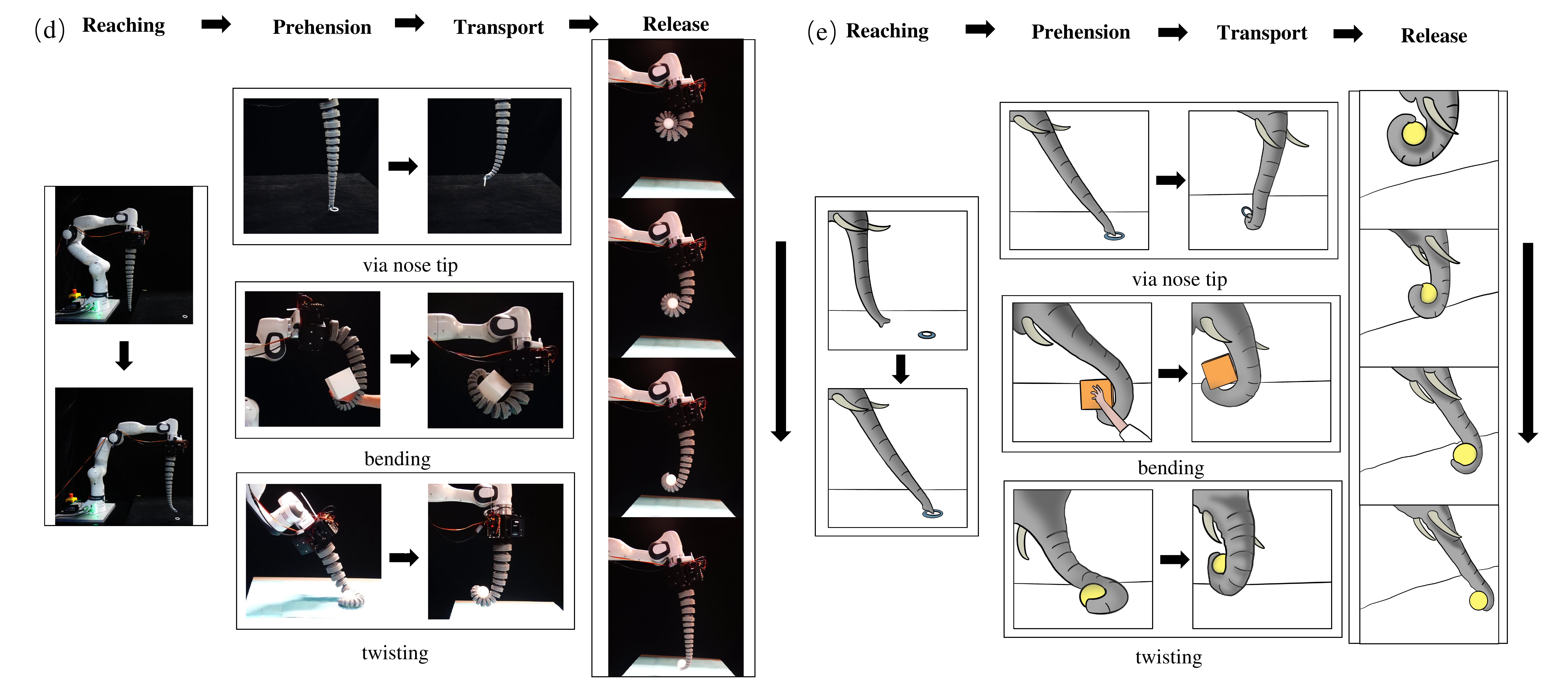}
\caption{The schematic graph demonstrates our hybrid system's capability to mimic the elephant motion primitives. (a).An overview of the proposed hybrid system compared to typical elephant motions. The system comprises a Franka Emika Panda robotic arm and a cable-driven soft continuum manipulator with a trunk-tip gripper. Three typical configurations are present: bending when lowering the head (middle), twisting with the head up (right), and resting under gravity (left). (b). The exploded CAD view of the motor housing demonstrates the detailed placement of the actuators. (c). The rotation of the steering gear creates a two-way open and close for the nose-tip gripper.
(d),(e) The schematic graph of the point-to-point elephant grasping procedure, including reaching, prehension, transport, and release phases. (The real trunk demonstration images are adapted from the video in \cite{dagenais2021elephants}). During the reaching phase, the primary deformation is elongation, which facilitates an extended reach. During prehension, small objects are grasped through an open-and-close motion of the \textcolor{blue}{trunk tip}, while larger objects exceeding the capacity of the nose tip are wrapped using the distal portion of the trunk, achieved through bending and/or twisting, depending on the object's position. In the transport phase, a curvature propagation would occur towards the proximal end to secure the object by inward bending. Finally, during the releasing phase, an opposite curvature propagation assists in releasing the object at the target location.}
\label{Fig1}
\end{figure*}

A typical point-to-point grasp by elephants can be decomposed into four main phases of motion: reaching, prehension, transport, and release \cite{dagenais2021elephants}, see in Fig. \ref{Fig1}(e). During the reaching phase, the elephant moves its head to adjust its orientation and elongates its trunk to fetch the target object. In the subsequent phases of prehension, transport, and release, the elephant trunk relies on three primary deformation methods---bending, twisting, and extension---to manipulate objects. The trunk bending, twisting, and extension are induced by curvature, torsion, and longitudinal strain, respectively, which are achieved through the activation of specific muscle groups. These motion primitives can be combined in different configurations to enable versatile grasping strategies. Through empirical observation of elephant grasping behavior in controlled experiments, 17 distinct strategies were summarized in \cite{dagenais2021elephants} for grasping objects of varying sizes, shapes, and weights by integrating these fundamental primitives with precise head orientation and movement.orientation and movement.

Inspired by these biological phenomena, this work presents a novel synergy between soft and rigid robotic systems to replicate the grasping strategies observed in elephants from a high level of behavioral perspective (Fig. \ref{Fig1}(a)).  The rigid component, a Franka Emika Panda robot arm, acts as an analog to the elephant's head, providing precise positioning and orientation within the workspace during the reaching phase. The soft manipulator in our hybrid system is designed to reproduce the motion primitives of an elephant trunk by a special actuation mode. By the cooperation of the synergy, the hybrid system is capable of replicating 9 grasping strategies out of the total 17 described, such as tip grip, distal wrap in the horizontal plane, or trunk "kick", with objects in various sizes and shapes.

It has been demonstrated that coordinating the degrees of freedom (DoFs) of the rigid skeleton with those of the soft trunk enhances the smoothness of overall shape planning \cite{morasso2024neural}. While there has been an initial attempt to study the theoretical framework of the synergy between rigid and soft robotic hydrostats — particularly from a cognitive and neural control perspective \cite{morasso2024neural, morasso2025computational}—the physical implementations remain scarce. Although there are a few existing studies that demonstrate physical implementations and offer insights from a control perspective \cite{uppalapati2020berry, peng2025dexterous, koe2025learning}, they fall short of providing high-level strategy-oriented approaches to coordination and manipulation. It is important to note that, in our context, a hybrid system refers specifically to a composition of a rigid skeleton coupled with a hyper-redundant continuum arm rather than an integrated structure of rigid and soft materials (e.g., bones and muscles) forming a single component.



In our soft manipulator, a symmetric three-cable configuration is specially arranged to achieve motion primitives in elephants inspired by their muscle group configuration \cite{longren2023dense}. One cable is positioned dorsally to mimic the dorsal longitudinal muscle of the trunk, inducing an upward bending motion when activated. The other two cables are placed ventrally to represent the symmetric ventral muscle groups. Equal activation of the ventral cables produces inward bending, while differential activation enables twisting. By actuating the respective cables in specific proportions, guided by our analytical model, the manipulator is capable of executing bending and twisting motions of varying shapes, mimicking the elephant trunk’s motion primitives.

By applying the geometrical concept of the logarithmic spiral to the design, bending motions can be propagated gradually from the distal to proximal sections, initiating at the tip and extending dynamically as more proximal regions curl. This behavior closely resembles the curvature propagation observed in an elephant trunk during prehension and transportation. Such dynamic bending is enabled by the inherent properties of the logarithmic spiral, wherein both joint geometry and corresponding bending stiffness decrease proportionally along the manipulator.

Inspired by the multifunctionality of the elephant trunk body and nose tip, we present a novel integration of a continuum manipulator and an end-effector gripper, each capable of independently executing grasping tasks. Current research often focuses on mounting grippers on soft or rigid robots, relying solely on the end-effector for grasping \cite{jiang2021hierarchical, gecko, xie2023octopus}, or exclusively using the soft arm for grasping \cite{wang2024spirobs, mcmahan2006field}. While these approaches achieve certain levels of functionality, they lack the synergistic integration of both elements, limiting their adaptability and efficiency in diverse tasks.

Most existing soft robotic arms are restricted to planar motions \cite{zhang2023situ, liu2019elephant, katzschmann2015autonomous}. Designs capable of achieving versatile configurations in three-dimensional (3D) space typically rely on multi-section architectures \cite{hannan2001elephant, zhou2022design, guan2020novel}, which require numerous actuators, significantly increasing control complexity. Alternatively, some single-section designs
\cite{kaczmarski2024minimal} exhibit rich configuration spaces but lack grasping capabilities, limiting their practical applications.


To enable diverse bending and twisting configurations in our soft manipulator, we develop a novel length-based forward kinematic model that leverages the mathematical properties of the logarithmic spiral. These kinematic primitives serve as the foundation for executing complex grasping tasks when combined with the rigid robot arm. While the logarithmic-spiral-based concept was first proposed in \cite{wang2024spirobs}, their friction-based model \cite{wang2024exploiting} was restricted to 2D motion, making it unsuitable for omnidirectional grasping. Traditional length-based modeling approaches, such as piecewise constant curvature methods \cite{webster2010design}, are not well-suited for cone-shaped robots. Adaptations for variable-curvature designs \cite{mahl2014variable} offer improved flexibility but are complex to derive.

Alternative modeling techniques, such as beam theories—including Cosserat rod theory \cite{alessi2024rod, alkayas2024soft}—and finite element methods (FEM) \cite{wu2022fem}, provide higher accuracy by analyzing forces and deformations. However, these methods demand significantly greater computational resources, making them less practical for real-time applications. In contrast, our approach balances mathematical simplicity with functional precision, enabling efficient modeling of the manipulator’s 3D configurations.


In contrast, our work adopts a different approach by developing a single-section cable-driven soft manipulator capable of full 3D motion and grasping. By integrating logarithmic spiral-based geometry and kinematic modeling, the proposed design overcomes the limitations of previous works, enabling a versatile configuration space and robust grasping capabilities. Combined with the rigid robot arm, the hybrid system achieves the precision, adaptability, and dexterity necessary to handle objects of varying sizes, shapes, and weights. These capabilities are demonstrated through a series of validation experiments.

\section{Results}
\label{sec:results}
Our contributions can be summarized as follows: 
\begin{enumerate}
    \item Drawing inspiration from the logarithmic spiral patterns observed in animal prehensile appendages, we identify a set of geometry constraint parameters that can be consistently applied to both a soft continuum manipulator and its end-effector gripper. This unified parameterization enables both components to achieve spiral-based grasping. Furthermore, we analyze the advantages of these spiral patterns in biological systems and demonstrate their relevance in robotic design.
    \item We establish a novel cable-driven actuation pattern for the soft manipulator to achieve bending and twisting, inspired by the functionality of the trunk muscles, which generate kinematic primitives. A novel length-based forward kinematic model is further proposed to describe the different configurations of these primitives, in which the model accuracy is validated via a set of experiments.
    \item We integrate the soft manipulator with a rigid robot arm to create a hybrid synergy that can achieve a wider range of configurations, effectively replicating 9 distinct grasping strategies observed in African elephant behaviors.
\end{enumerate}

In this section, we first present an overview of the design and modeling of our robotic system in Section \ref{subsec: design_and_modelling}, followed by an overview and validation of our forward kinematic model in Section \ref{subsec: K-model}. The reproduction of  9 grasping strategies inspired by the natural behavior of elephants is shown in Section \ref{subsec: real_trunk_eval}.
More technical details of our work are postponed to Section \ref{method}. Visualizations of our system and experiments are shown in the supplementary video.

\begin{figure*}[htbp]
    \centering
    \includegraphics[page=1, width=\textwidth]{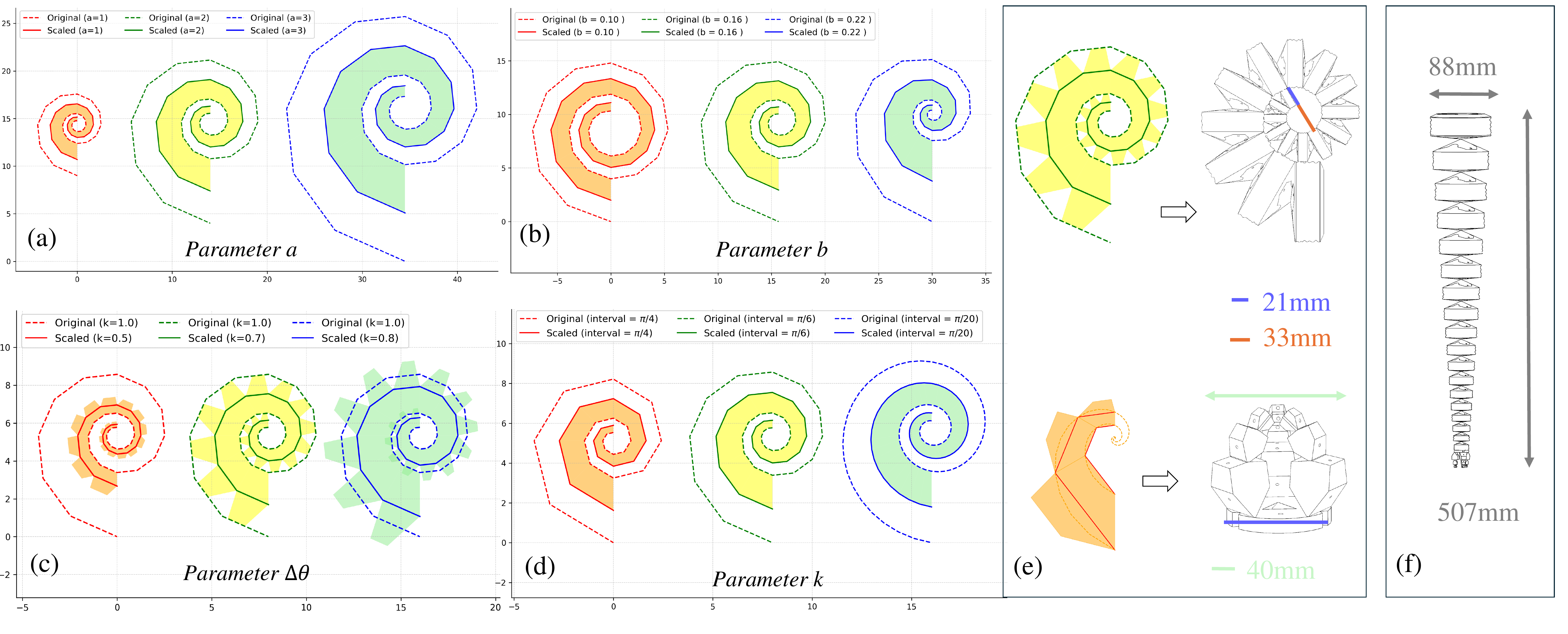}\\
    \includegraphics[page=1, width=\textwidth]{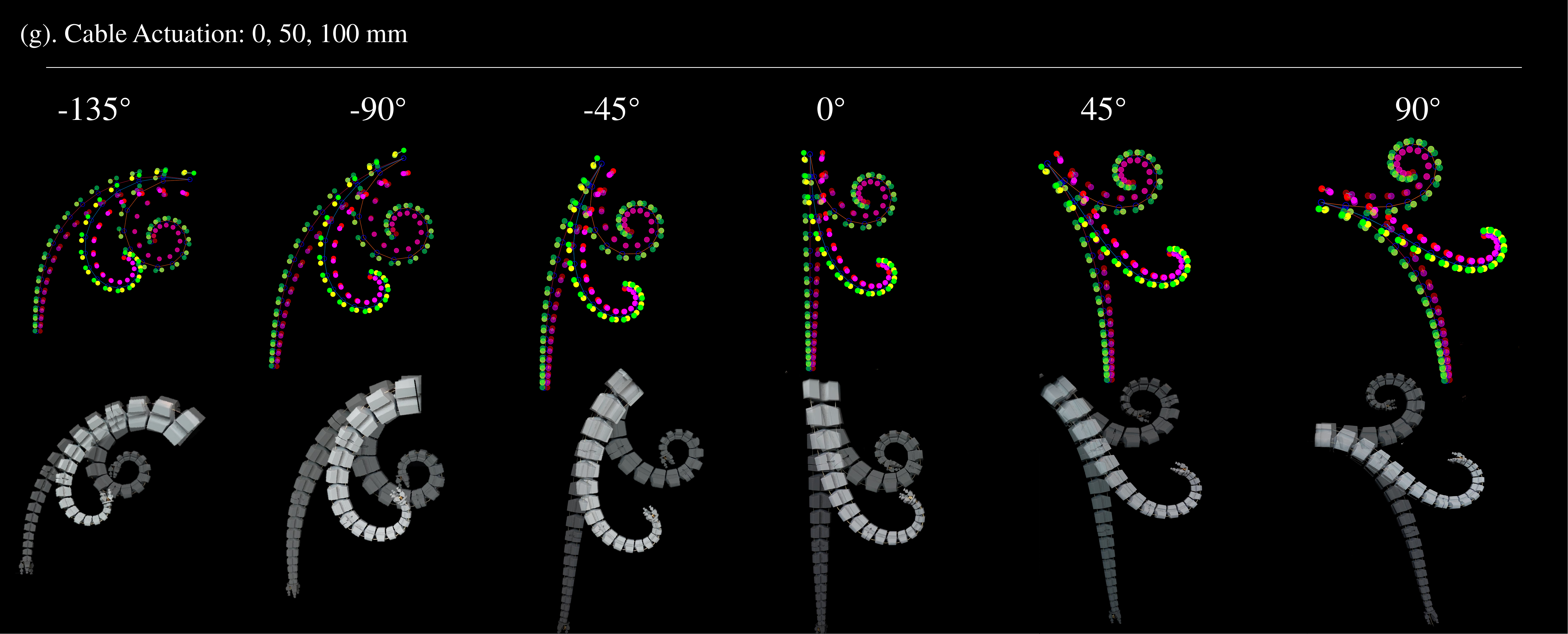}\hfill
    \includegraphics[page=1, width=\textwidth]{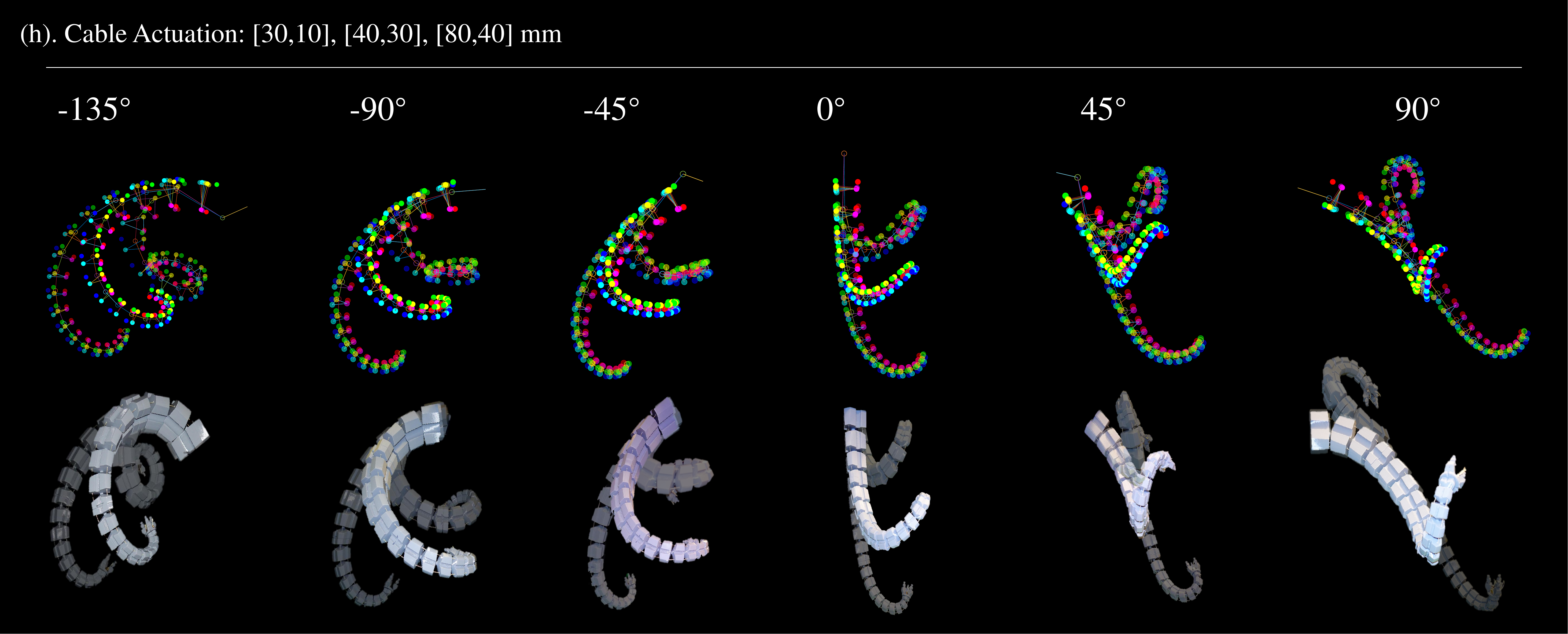}\\
    \caption{(a). Three sprials at different scales formed by varying the scale factor $a$. (b). By varying the growth parameter $b$, we can change the growth rate of the spiral. (c). The compactness parameter $k$. (d). The discrete parameter $\Delta\theta$. (e). The optimal parameters are selected for the geometry constraint of our soft manipulator and its end-effector gripper. (f). The final dimensions of our soft manipulator are determined based on the selected optimal parameters. (g),(h): The qualitative comparison of bending and twisting between the analytical model (above) and prototype (below) under the same actuation amount in different gravity directions.}
    \label{fig:design_model}
\end{figure*}

\begin{figure*}[htbp]
    \centering
    \includegraphics[page=1, width=\textwidth]{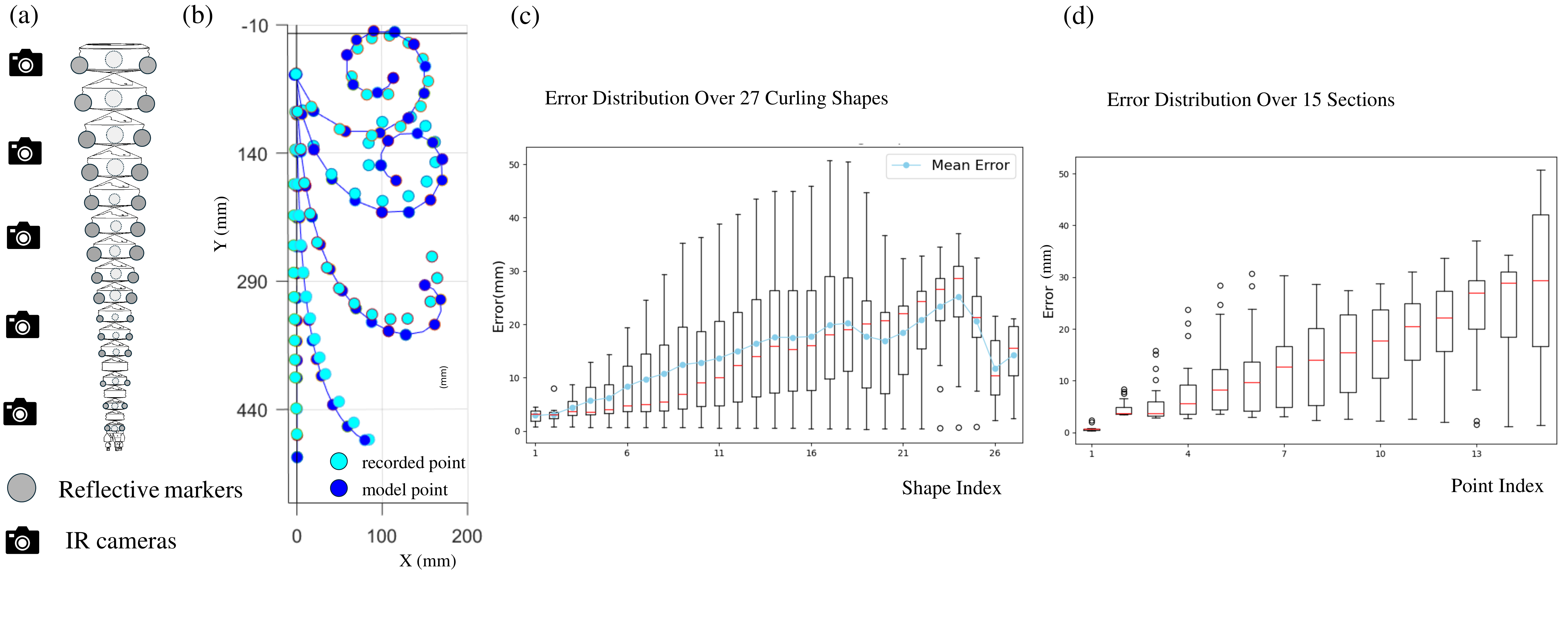}\\
    \includegraphics[page=1, width=\textwidth]{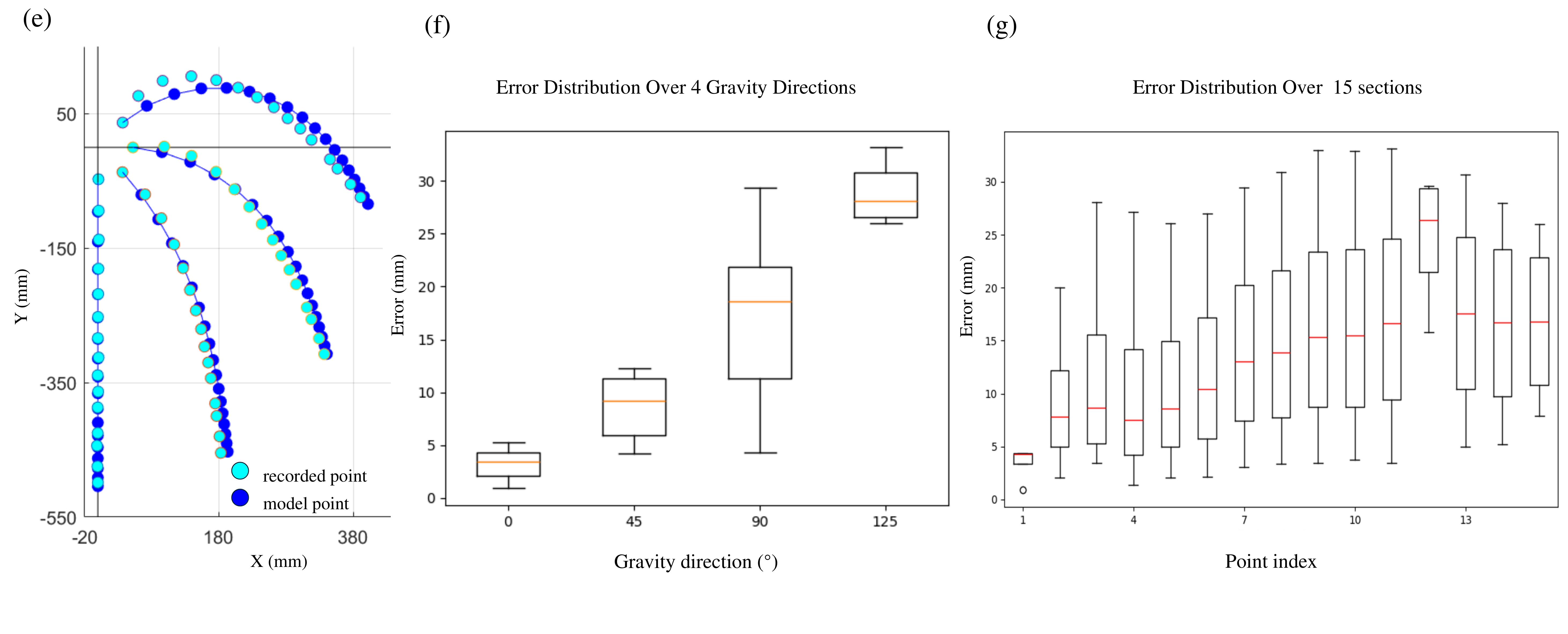}\   
    \caption{(a). Schematic graph of the motion capture validation experiment for the model. (b). The visualization of the motion capture results in the vertical plane for different bending shapes, and the quantitative result shows the averaged error along shapes (c) and sections (d). (e). The visualization of the motion capture results in the resting shapes under different gravity angles, and the quantitative result shows the averaged error along shapes (f) and sections (g).}
    \label{fig:validation}
\end{figure*}

\subsection{Soft Manipulator Design}
\label{subsec: design_and_modelling}
In this section, we first explain how to determine optimal parameters in the logarithmic spiral and then transform the obtained spiral curve into the real prototype. We describe the logarithmic spiral in the Cartesian coordinate as:


\begin{equation}
\begin{cases} 
x = -ae^{b\theta} \cdot \cos\theta \\
y = ae^{b\theta} \cdot \sin\theta
\end{cases}
\end{equation}
As $\theta$ increases from negative to positive infinity, the curve will spiral outward from the origin in cycles. We position two logarithmic spirals with the same parameters together and scale one of them by scaling factor $k$:

\begin{equation}
\begin{cases} 
x_k = -k \cdot x \\
y_k = k\cdot y
\end{cases}
\end{equation}

These two contour spirals would form an enclosed area, as shown in Fig. \ref{fig:design_model}(a). The colored area enclosed by the two contour spirals determines half of the main body of our manipulator, and a rotation around one contour spiral constitutes the whole manipulator as shown in Fig. \ref{fig:design_model} (e). The body shapes vary under different parameters. First, we list three design principles to determine the optimal parameters:

\begin{principle}
The design shall keep compact.
\label{p1}
\end{principle}

\begin{principle}
The grasping range of the trunk body and nose-tip gripper shall be overlapped to eliminate ungraspable sizes.
\label{p2}
\end{principle}

\begin{principle}
The soft prototype shall match the rigid robot arm.
\label{p3}
\end{principle}

As illustrated in Fig. \ref{fig:design_model}, the key parameters exert different influences on the resulting spiral shapes. A brief overview is provided here, while detailed derivations are present in Section \ref{method}.

Fig. \ref{fig:design_model}(a) illustrates how $a$ determines the scale of the spiral, which would subsequently decide the manipulator length and size. When $a$ increases, the overall size of the spiral increases. A proper $a$ can be decided via Principle \ref{p3}.

Fig. \ref{fig:design_model}(b) illustrates how $b$ affects the growth rate of the spiral. The larger the value of $b$, the faster the curve decays towards the origin (Fig. \ref{fig:design_model}(b)). If the full length of curve is kept constant, a reduction in $b$ leads to a smaller area in the center of the spiral (drawn in orange line in Fig.  \ref{fig:design_model} (e)), with a smaller width of the end section (drawn in blue line in Fig.  \ref{fig:design_model}(e)). The width of the smallest section imposes a constraint on the maximum allowable size of the mounted end-effector gripper, while the central encirclement area of the spiral determines the minimum size of a graspable object of the trunk body. Thus, a larger $b$ reduces the grasping range of the end-gripper while benefiting that of the trunk manipulator. A tradeoff of $b$ to fulfill Principle \ref{p2} is necessary to improve the overall graspable range and reduce the difficulty of small-scale fabrication.

$k$ determines the degree of compactness of the spiral. An appropriate value of $k$ (the yellow spiral in Fig. \ref{fig:design_model}(c)) enables the spiral to curl tightly in successive loops without gaps, which is crucial for Principle \ref{p1}.

$\Delta \theta$ defines the amount of discretization of the spiral. We compare different values of $\Delta\theta$ in Fig. \ref{fig:design_model}(d). The discrete level is restricted by the trade-off of the prototype dexterity and the fabrication difficulty. A more continuous prototype would be more flexible yet more difficult to fabricate. Empirically, the $\Delta\theta$ has been determined as $\frac{\pi}{6}$ for the trunk body. 

The spiral geometry for the tip gripper exhibits a higher degree of angular discretion compared to the trunk-body spiral, as shown in Fig. \ref{fig:design_model}(e). While the tip gripper is designed to be mounted at the distal end of the trunk manipulator, its design inspiration is drawn not from the elephant trunk but from human fingers. Prior studies \cite{gupta1998motion, hutchison2010fibonacci, littler1973adaptability} have demonstrated strong correlations between human fingers and logarithmic spirals, both in terms of anatomical structure and motion trajectories. To maintain design consistency, we therefore adopt human fingers as the reference model for the gripper. Specifically, the spiral for the finger design is constructed using the golden ratio, $b = 0.618$ \cite{hutchison2010fibonacci}, and a larger angular discretization of $\Delta \theta = 60^\circ$.

Fig. \ref{fig:design_model}(e) demonstrates how we transform the spiral geometry to the prototype. Additionally, each section of the manipulator is wrapped with wrinkled silicon shells. The ratio of the sizes of the adjacent sections remains constant ($k_p = 0.9196$). This proportion lays the foundation for developing the forward kinematic model. According to \cite{wang2024spirobs}, the graspable range of the trunk main body spans objects with cross-sectional widths between $267 mm$ and $33 mm$. The maximum theoretical payload of $4.89 kg$ is achieved when grasping objects with a cross-sectional width of $33 mm$. The tip gripper is capable of grasping objects with a maximum cross-sectional width of approximately $40 mm$, resulting in a $7 mm$ overlapping range with the trunk body. It should be noted that this payload estimation considers only the soft manipulator. In practice, the overall payload capacity of the hybrid system may be further constrained by the payload limits of the supporting robot arm.

\subsection{Forward Kinematic Model}
\label{subsec: K-model}
This section provides an overview of the forward kinematic model, with detailed derivations deferred to Section \ref{kinematic}.

The manipulator’s elastic joints are designed as cylindrical components with radii that decrease progressively according to a fixed scaling ratio $k_p = 0.9196$. This proportional reduction in joint geometry leads to a corresponding distribution of decreasing bending stiffness along the manipulator. As a result, when the actuation cables are tightened, the curling motion initiates at the tip and propagates toward the base. To establish the forward kinematic model, we make the following two assumptions:


\begin{assumption}
Each link of the manipulator is treated as a rigid body.
\end{assumption}\label{as1}

\begin{assumption}
The joints are assumed to allow bending deformation, while torsional effects are neglected.
\end{assumption}\label{as2}

The joint dimensions are significantly smaller than the link radius, allowing the manipulator to be modeled by a joint-link configuration similar to that of rigid robots. When the cable is actuated, while the links are undeformed, the joints rotate due to the large passive material deformations. These deformations are constrained by the geometric of the links, which impose rotation limits of $30 ^{\circ}$ at each joint.





\subsubsection{Resting under gravity}
We start with analyzing the initial resting state of the manipulator. In the absence of actuation, if the manipulator base adopts an initial inclination angle, each section would bend under the influence of gravity. The material stiffness of the joints provides resistance to this bending, resulting in a shape model that describes the manipulator's configuration at various inclination angles. This static shape serves as the baseline for further modeling under cable actuation. Notably, the influence of gravity allows the soft manipulator to achieve a more diverse range of configurations. When gravity is considered, the same actuation input induces different shape configurations depending on the orientation of the manipulator's base relative to the gravitational field. This highlights the importance of incorporating gravitational effects into the model to capture the manipulator's true behavior and expand its functional range.

\subsubsection{Bending}

Bending deformation is affected by the cable actuation amount, the manipulator’s geometry and material properties. When a single cable is shortened by a specified amount, the resulting length change is distributed across all sections of the manipulator, inducing corresponding joint rotations. To accommodate the deformation, the other two cables must be passively relaxed. By analyzing the system’s geometry, material stiffness, and torque balance, the relationships governing length distribution between adjacent sections and the associated rotation angles can be calculated. The same methodology applies when both ventral-side cables are equally shortened while the remaining dorsal cable is relaxed, resulting in symmetrical bending of the manipulator. This approach enables accurate prediction of the manipulator’s shape in response to a given actuation input. Detailed derivations are described in Section \ref{kinematic}.

\subsubsection{Twisting}
Twisting deformation is introduced by applying different amount of actuation in the second cable compared to the actuation of the first cable. To model this behavior, we propose the following assumption:

\begin{assumption} \label{as3}
The actuation of the second active cable does not alter the distribution relationships established by the initial cable.
\end{assumption}

Under this assumption, the second cable’s action is treated as a further rotation of the joint around a specific axis defined by the initial cable’s actuation amount. This approach enables the calculation of joint angles under two-cable actuation, capturing the twisting effect induced by differential cable tensions. The section geometry imposes constraints on the manipulator’s deformation by limiting the maximum bending angle to $\theta = 30^\circ$, while the allowable twisting angle is depending on the state of the first cable. If the distributed cable length causes a joint to reach its limit, the excess cable length is reallocated to subsequent joints. This redistribution ensures smooth and continuous deformation while respecting geometric and material constraints. Detailed procedures for handling these limits are described in Algorithm \ref{algo1} and Algorithm \ref{algo2}, presented in Section \ref{kinematic}.





\subsubsection{Model Validation}

We evaluate the manipulator’s performance in replicating predefined shapes predicted by the forward kinematic model. We utilize an OptiTrack motion capture system equipped with 17 cameras to track the manipulator's shape changes under different cable actuation and gravity directions. The evaluation process involves quantitative and qualitative assessments of the manipulator's ability to conform to target geometries. 

The validation focused on two behaviors central to elephant trunk motion: bending and resting under gravity. Three reflective markers are affixed to the lower edge of each section, forming rigid body coordinate frames along the manipulator’s backbone, as shown in Fig. \ref{fig:validation} (a). Actuation commands were sent to the GM6020 motors via the RoboMaster Development Board to selectively tighten or relax the cables embedded in the manipulator. Once the manipulator reached static equilibrium, its shape was recorded, and the 3D positions of the rigid body frames were stored and aligned with the model output using a rigid-body transformation. The position deviation was calculated at each frame of the backbone, and the Root Mean Square Error (RMSE) was used as the primary accuracy metric. The detailed systematic evaluation is described in Section \ref{method_mocap}.\\
\textbf{Bending:} We evaluate the entire bending process from the initial state---when the manipulator is fully extended--- to the fully wrapped state in the vertical gravity direction. The qualitative and quantitative results are shown in Fig. \ref{fig:design_model}(g) and Fig. \ref{fig:validation}(b), respectively. In Fig. \ref{fig:validation} (b), we select five frames among the 26 discrete bending steps. Quantitatively, the average RMSE value for all sections across all steps was $10.41 mm$, with the largest deviation being $32.14 mm$. Fig. \ref{fig:validation}(c) shows how the absolute error distribution changes as the shape of the manipulator changes with each step. Fig. \ref{fig:design_model}(d) shows the absolute error distribution of each section across the 27 shapes captured by the motion capture system.\\
\textbf{Deformation under gravity:}
To validate the accuracy of the gravity model, we allow the manipulator to hang naturally at tilt angles of $0^\circ$, $45^\circ$, $90^\circ$, $135^\circ $, with respect to the gravity direction. The qualitative and quantitative results are shown in Fig. \ref{fig:design_model}(h) and Fig. \ref{fig:validation}(e), respectively.
Quantitative analysis revealed RMSE values of $5.48mm$, $10.57mm$, $10.45mm$, and $19.86mm$ for the respective tilt angles. The positional error distributions are further illustrated using box plots in Fig. \ref{fig:validation}(f) and \ref{fig:validation}(g), offering insights from directional and sectional perspectives, respectively. These results demonstrate the model's effectiveness at smaller tilt angles, with increasing deviations observed at larger angles due to cumulative material deformation and gravitational effects.

The results demonstrated a fairly consistent match between the experimental and theoretical profiles. The minor deviations observed could be attributed to slight variations in the material's elastic properties and random errors introduced during the fabrication process. Further details about the potential sources of error that could affect the accuracy of our model validation results are discussed in Section \ref{subsec: SoE}. These findings validate the manipulator's capacity to mimic bending behaviors, showcasing its potential for adaptive and gravity-compliant tasks.



\subsection{Reproducing Elephant Grasping Strategies with the Hybrid System}
\label{subsec: real_trunk_eval}

\begin{figure*}[htbp]
\centering
\includegraphics[width=\textwidth]{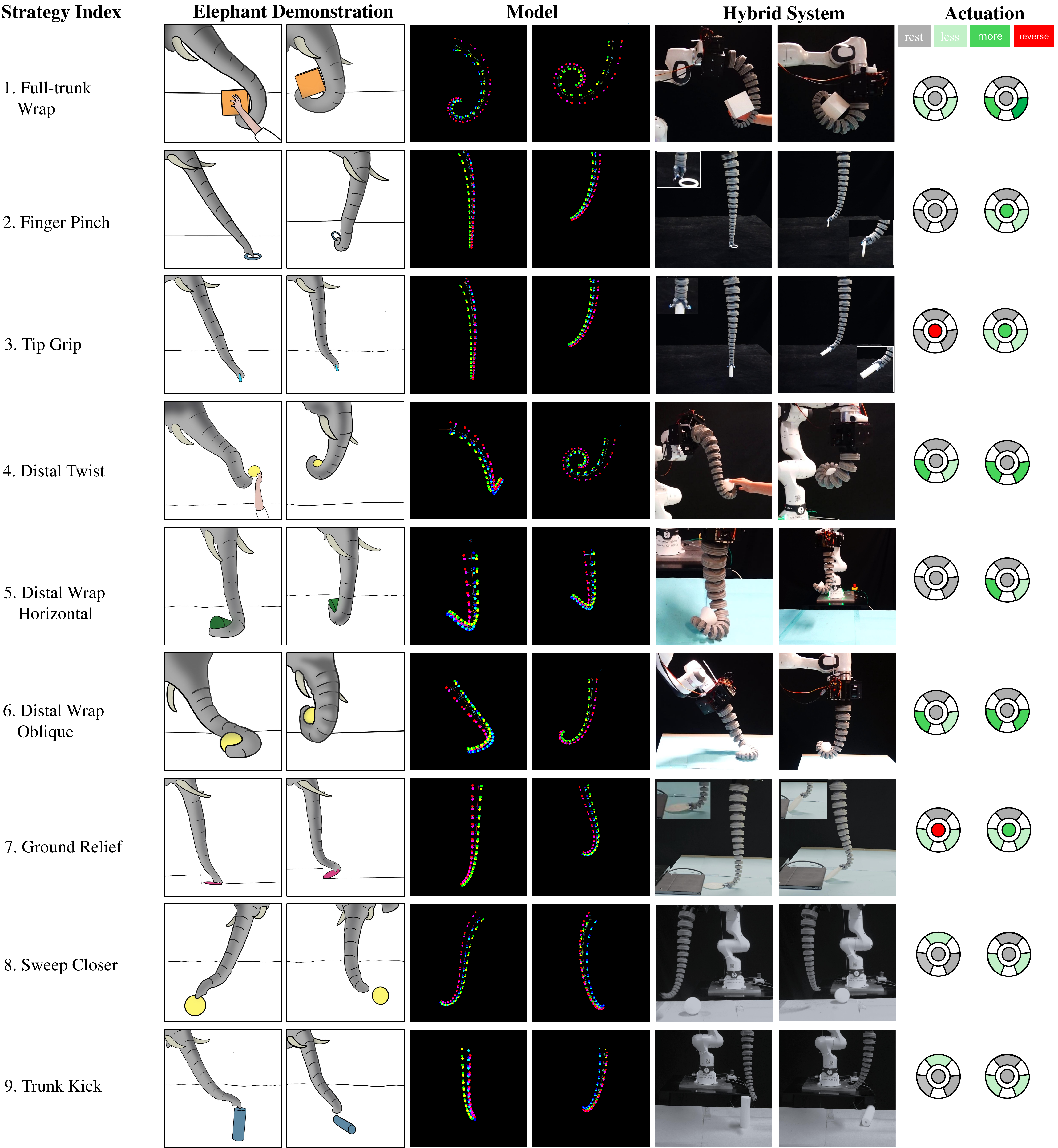} 
\caption{Demonstration of the hybrid system's capability of reproducing the 9 distinct elephant grasping strategies. The listed strategy and elephant demonstration are derived from studies in \cite{dagenais2021elephants}. Key shapes with different actuation patterns are determined from the study and pre-defined by the forward kinematic model. The snapshots of our prototype demonstration showcases the ability of a successful prehension and transportation of geometric objects. Full video demonstration can be seen in the supplementary files.}
\label{grasp_strategies}
\end{figure*}

To evaluate the capability of our hybrid robotic system to reproduce the grasping strategies of elephants, we conducted a series of experiments inspired by the behaviors documented by Dagenais et al. \cite{dagenais2021elephants}. Our experiments utilized the same geometric objects as those used by Dagenais et al for real elephants. These objects provided a standardized basis for comparing the performance of our robotic system with that of a real elephant. Objects were placed at various distances and orientations within the robot’s workspace to test the system’s ability to perform diverse prehension tasks.


Out of the 17 grasping strategies demonstrated by the real elephant in \cite{dagenais2021elephants}, our hybrid system successfully replicated 9 of the strategies. Fig.\ref{grasp_strategies} summarizes the qualitative results of our experiments and also the actuation methodology utilized to form the shape required to execute the grasping strategy. We refer the reader to our supplementary video for a complete demonstration of the grasping strategies.

\subsubsection{Strategy Breakdown}

If we break down the elephant strategy in Fig. \ref{Fig1} further, we find that the variations primarily occur during the prehension stage, which is also the most complex of the four steps. Broadly speaking, the prehension phase involves a preparatory action,  followed by one of two distinct modes, depending on the characteristics of the target object: nose-tip prehension or trunk-body prehension.

\begin{enumerate}
    \item \textbf{Preparatory Action}: When objects are positioned beyond direct reach or oriented in a manner unsuitable for immediate grasping, elephants typically perform a preparatory action to reposition the item. A trunk ‘kick’ or sweeping motion (strategies 8 and 9 in Fig. \ref{grasp_strategies}) is usually executed to roll or push the object toward a more accessible location. This strategy is commonly employed for small or lightweight objects and requires fine control of the distal trunk and nose tip to maintain precision. Additionally, elephants would also press objects against the ground with their trunk to stabilize them before grasping. 
    \item \textbf{Nose Tip Prehension:} For small and light objects, elephants utilize the nose tip to perform pinching, gripping and/or suction, with the direction from the top or the side determined by the target pose. Pinching (strategy 2 in Fig. \ref{grasp_strategies}) is typically for thin and slender objects, involving minimal contact—analogous to the use of fingertips in humans. Gripping (strategy 3 in Fig. \ref{grasp_strategies}) is used for objects with similar shapes but slightly larger sizes, where the entire “palm” of the trunk wraps tightly around the object. In the case of granular materials (e.g., a pile of peanuts) or smooth, flat objects (e.g., a piece of glass), elephants often combine suction with gripping to facilitate effective prehension.
    \item \textbf{Trunk Body:} When interacting with larger, heavier, or irregularly shaped objects, elephants engage the entire trunk body during prehension. The strategy employed is determined by the position of the object, whether it is suspended or resting on a surface. In the case of suspended objects, bending serves as the primary deformation mode. During bending (strategy 1 in Fig. \ref{grasp_strategies}), the distal portion of the trunk wraps around the object within the vertical plane, ensuring a secure grasp. This method is particularly effective for cylindrical or bulky items. 
    Conversely, when the object's base is in contact with the ground, twisting becomes essential (strategies 4, 5, and 6 in Fig. \ref{grasp_strategies}). A purely upward or inward bend would induce trunk encirclement in the vertical plane, which may impede effective grasping due to interference with the ground. By applying a twist, the curvature direction gradually transitions from the vertical plane to the oblique plane and finally to the horizontal plane, enabling the trunk to encircle the object from the side. This strategy is particularly effective for objects with a broad or flat base situated on a surface.  
    In addition, elephants occasionally rely on their tusks or forelimbs to aid in the manipulation and handling of particularly large or heavy items.
\end{enumerate}



During the transport phase, bending remains the dominant deformation mode. The inward curvature propagates further from the distal end to the proximal sections, mimicking the behavior of an elephant trunk. For nose-tip prehension mode, a distal flip is frequently used to secure the object after lifting. This facilitates secure transportation by engaging larger volumes of the manipulator for stabilization and enhancing grip strength for heavier or bulkier objects. 

Throughout the prehension and transport processes, elephants modulate their trunk stiffness to accommodate varying object loads. This is achieved via localized multi-axial muscular contractions. Similarly, our soft manipulator modulates stiffness through tendon antagonism. The manipulator can maintain its shape by dynamically adjusting opposing tensions in the dorsal and ventral cables while achieving secure prehension and stable transportation. However, the stiffness regulation can only be executed in the sagittal plane since the longitudinal cables could not resist torsional disturbance. This limitation is further discussed in the future work in Section \ref{future} below.




During the releasing phase, a curvature propagation in the opposite direction occurs, transporting the object from the proximal end toward the nose tip, and finally let go to the desired target point.

\subsubsection{Results and Observations}
The experiments demonstrate the ability of the soft manipulator to reproduce the motion primitives of an elephant trunk effectively. Fig. \ref{grasp_strategies} highlights the qualitative success of each replicated strategy, while the actuation patterns show the versatility of the manipulator’s cable-driven mechanism. Notably, the logarithmic spiral design enabled dynamic bending propagation, ensuring the manipulator could adapt its shape in response to the object's shape and position.

\section{Discussion}
\label{sec:discussion}

\subsection{Advantages and Limitations of Logarithmic Spiral Geometry in Robotic Design}
The logarithmic spiral serves as an effective design reference, as it is a recurring pattern observed not only in elephants but also across various prehensile appendages in different species. This pattern is evident in anatomical structures \cite{hutchison2010fibonacci}, motion trajectories \cite{gupta1998motion}, and muscle morphologies \cite{wang2024spirobs}. Combined with its mathematical properties and observed behaviors in nature, we hypothesize that the logarithmic spiral offers significant advantages for prehension, both biologically and in engineered systems.

From an engineering perspective, the logarithmic spiral contributes to enhanced grasping ranges while simplifying the design, fabrication, and modeling of robotic systems. First, the self-similarity inherent to the logarithmic spiral makes it a scalable design principle, applicable across different size scales. Prior research \cite{wang2024spirobs} has demonstrated that grasping capabilities remain consistent across meter-scale to millimeter-scale designs, highlighting its cross-scale adaptability. Second, the tapering geometry constrained by the logarithmic spiral supports a broader curvature range, improving grasping adaptability. This is because a uniform activation in longitudinal muscles (or equivalent actuation mechanisms) leads to increased local curvature as the cross-sectional radius decreases \cite{kaczmarski2024minimal}. In other words, the more pronounced the tapering, the greater the grasping precision and adaptability. Third, the fixed ratio of geometric dimensions from proximal to distal sections unifies the design, enabling modular fabrication. A single-unit design can be scaled up or down systematically for adjacent sections, reducing complexity in manufacturing while maintaining functionality.

However, there are limitations to the application of the logarithmic spiral in robotic systems. This study primarily focuses on replicating grasping behaviors observed in elephants, while other trunk movement patterns fall outside the scope of this work. For instance, elephants exhibit behaviors such as forming pseudo-joints for sideways reaching or actuating only distal segments while keeping proximal sections rigid. These movements involve complex biomechanical strategies that extend beyond the capabilities of a logarithmic spiral-based manipulator. Consequently, while the logarithmic spiral serves as a robust framework for grasping, it represents only a subset of the full behavioral repertoire of elephant trunk movements.

\subsection{Performance Evaluation and Future Work}\label{future}
The results presented in this study demonstrate the effectiveness of the proposed soft-rigid hybrid robotic system in replicating key grasping strategies inspired by the elephant. The system successfully reproduced 9 out of the 17 documented strategies, showcasing its ability to adapt to objects of varying shapes, sizes, and orientations. This work showcases the first implementation of a logarithmic-spiral-based soft manipulator integrated with a rigid robotic arm, effectively leveraging the strengths of both components. The rigid arm provides precise positioning and support, while the cable-driven soft manipulator achieves complex motion primitives such as bending, twisting, and tip grasping. The experimental validation highlights the versatility of the system and underscores the advantages of the logarithmic spiral geometry in enabling dynamic bending propagation and controlling complex manipulator shapes with minimal actuation inputs. The forward kinematic model developed for the manipulator further contributed to this achievement by enabling precise control of the manipulator's motion primitives, validated through experimental results.

While our experiments demonstrate the manipulator’s ability to perform predefined grasping strategies, the system lacks adaptability to novel objects and dynamic tasks. This is in part due to the reliance on predefined actuation models in this work rather than on a more flexible learning-based framework.

There were also three types of strategies that our system could not replicate, which reveal several other design limitations:
\begin{enumerate}
    \item \textbf{Suction Capability:} Elephants often utilize suction to grasp granular, smooth, or flat objects, which is beyond the current capabilities of our system as it lacks mechanisms for generating or controlling suction forces.
    \item \textbf{Pressing objects down on the ground:} Elephants employ strategies where they press objects against the ground to stabilize them before grasping, particularly for irregularly shaped items. Our actuation system is designed based on position control rather than force control. This approach simplifies our control methodology, but at the same time makes it difficult to precisely regulate the interaction forces between the manipulator and the object. For example, pressing an object down using our manipulator would require the relationship between the torque on the motors (to maintain the manipulator's stiffness) and the force exerted vertically on the object by the manipulator to be known. However, this relationship cannot be derived from our current analytical model.
    \item \textbf{Use of other body parts to assist in grasping:} Elephants occasionally rely on their tusks or forelimbs to aid in manipulation and grasping very large or heavy objects, an aspect that our single manipulator and rigid arm configuration cannot replicate.
\end{enumerate}

We envision several potential directions for our future work. Firstly, we could improve stiffness control and mitigate torsional vulnerabilities by exploring alternative cable actuation configurations. Secondly, incorporating force control mechanisms will enable more precise grasping force regulation and improve adaptability to object size and material variations. To advance this work even further, a more intelligent grasping system based on a deep learning framework could be developed and tested in unstructured environments involving irregularly shaped objects and dynamic disturbances. These scenarios will provide deeper insights into our system's robustness and applicability in real-world tasks.


 

\section{Methods}
\label{method}
In this section, we delve into the details of the method and experiments. Firstly, we provide the method to derive the key parameters in the logarithmic spiral in design (Section \ref{spiral_para}). Next, we introduce the fabrication details of the hybrid system (Section \ref{method_hardware}), followed by a thorough derivation of the forward kinematic model (Section \ref{kinematic}). Lastly, we illustrate the procedure for the motion capture experiment (Section \ref{method_mocap}).

\begin{figure*}
\centering
\includegraphics[width=\textwidth]{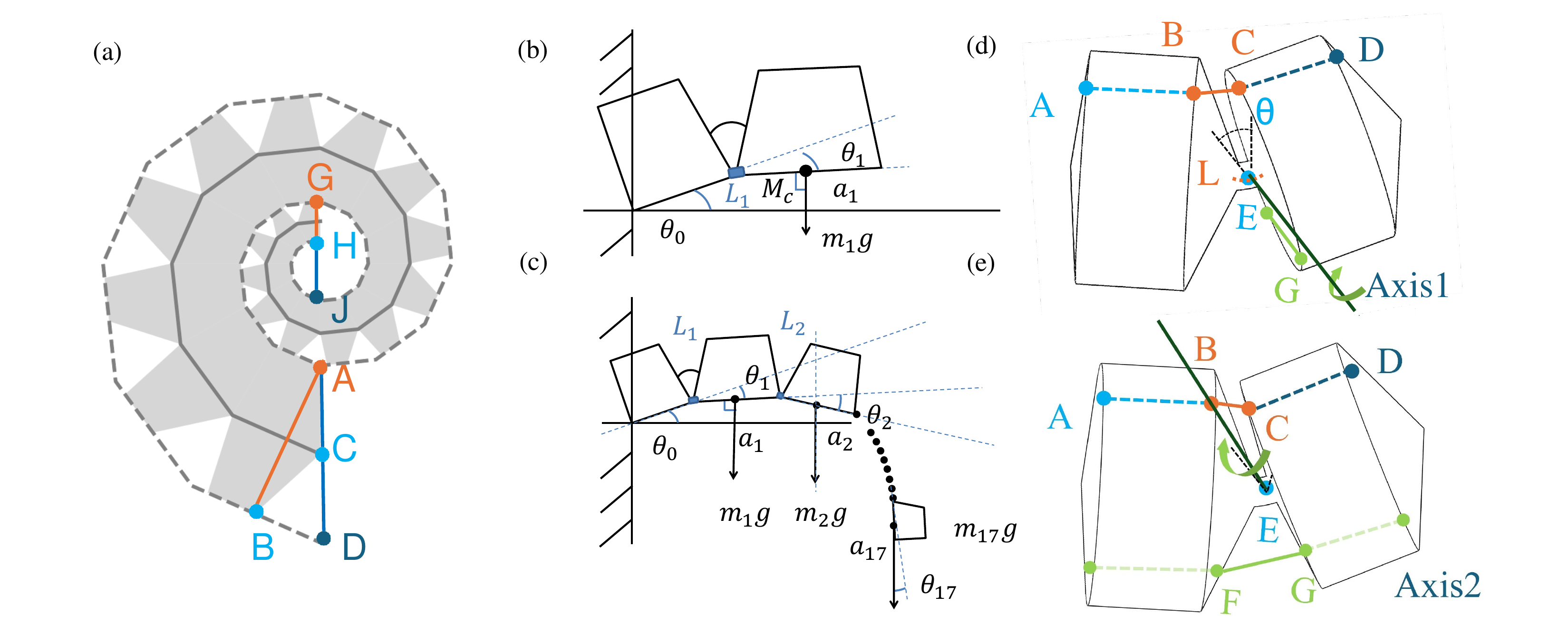} 
\caption{(a). Schematic graph of the spiral with optimal parameters. (b). A two-section arm hangs freely with an incline angle $\theta_0$ under gravity. Notably, only the upper half of the body is shown here and in (c) for simplification. (c). The generalization of force equilibrium to multiple sections. (d). Bending deformation. The points $A, B, C, D, E, G$ represent the positions of the holes where the cables run through, while the colored lines represent cables. $Axis 1$ is obtained by transforming the vector $EG$ to pass through point $E$. When tightening cable $ABCD$, the section on the right would rotate around $Axis 1$. When a series of sections rotate around their own axis (which is parallel with each other), a bending effect is induced where all the joint rotation angles are in the same plane. (e). Twisting deformation. On top of the bending effect, if the second cable $FG$ is subsequently tightened, in order to maintain the length of $ABCD$, the section on the right would rotate along $Axis 2$, which is defined by the vector $BE$. When a series of sections rotate around their own axis (which is not parallel with each other), a twist is induced where the joint rotation angles would not be in the same plane.}
\label{fig4}
\end{figure*}

\subsection{Determining the Optimal Parameters of the Logarithmic Spiral}
\label{spiral_para}


The starting and ending angle of the spiral contours is set as $\theta \in  (\frac{\pi}{2},\frac{7\pi}{2})
$, with the discrete angle $\Delta\theta = 30^\circ$.
The radius of the base section of the trunk manipulator $R_{soft} = AB$ shall match the radius of the last link of the rigid robot arm $R_{rigid}$ as shown in Fig.\ref{fig4} (a). From Fig.\ref{fig4} (a), it can be derived that:

\begin{equation}
    a = \frac{R_{rigid}}{R_{soft}}
\end{equation}

To keep the spiral compact, the outer contour of the trunk body shall fit closely with each successive loop when curled. The most compact shape is achieved when the structure aligns precisely between the inner and outer loops of the original spiral, where $AC = CD$, as shown in Fig. \ref{fig4} (a). According to the geometry, let the $y$ coordinate of $PointA$ be $y_\text{inside}$, $PointD$'s as $y_\text{outside}$, while $PointC$'s as $y_\text{k}$, $k$ can be expressed as follows:

\begin{align}
        y_k = \frac{y_{\text{inside}} + y_{\text{outside}}}{2}
        \label{equ2}
\end{align}

\begin{align}
\begin{cases}
y_{\text{inside}} &= ae^{b(\theta - 2\pi)} \sin(\theta - 2\pi) \\
y_{\text{outside}}& = ae^{b\theta} \sin\theta \\
y_k& = kae^{b\theta} \sin\theta
\label{equ3}
\end{cases}
\end{align}


    
The tip gripper is responsible for grasping small objects. To satisfy the requirement in Principle \ref{p2}, the maximum effective graspable size of the tip gripper shall be larger than the minimum graspable size of the trunk main body, which means their grasping ranges shall overlap. This is determined by the value of parameter $b$ and the finger geometry.

Referring to Fig. \ref{fig4} (a), the width of the tip gripper base is $GH$, which is equal to the diameter of the robotic arm's end section. The minimum graspable size for the soft arm main body is the distance between the two points in the inner circle $HJ$. If the maximum effective graspable size of the tip gripper is $m$ times bigger than the gripper base, we establish that:

\begin{align}\label{eq77}
HJ &= \sqrt{(x^2_{\theta_{\text{min}}} - x^2_{\theta_{\text{min}}+\pi})^2 + (y^2_{\theta_{\text{min}}} - y^2_{\theta_{\text{min}}+\pi})^2} \\ \nonumber
&= m \cdot HJ
\end{align}

\begin{align}\label{eq88}
GH = \sqrt{(x^2_{\theta_{\text{min}}} - x^2_{\theta_{\text{min}}+2\pi})^2 + (y^2_{\theta_{\text{min}}} - y^2_{\theta_{\text{min}}+2\pi})^2}
\end{align}

Analyzing the finger geometry, we empirically determined the $m = 1.53$. Combining equation \ref{equ3},\ref{eq77},\ref{eq88}, we get:

\begin{equation}\label{eq10}
\begin{cases} 
k = \frac{1}{2} + \frac{1}{2} \cdot e^{-2b\pi}\\
\frac{1}{2(1-k)}\left(e^{-2\pi b} + e^{-\pi b}\right) = m
\end{cases} 
\end{equation}


Based on these three principles, the parameters of the logarithmic spiral were uniquely determined to constrain the robot geometry, which provides support for the subsequent fabrication and modeling.

\subsection{Hardware Fabrication}\label{method_hardware}

As shown in figure \ref{Fig1} (a), the hybrid system consists of a Franka Emika Panda robot arm, a motor housing, a spiral-based soft manipulator, a set of wrinkled silicone shells, and a tip gripper with three fingers. 


Based on the spiral parameters determined in subsection \ref{spiral_para}, the enclosed quadrilateral area between two contour spirals can be rotated around the central axis to form each section of the soft arm main body. 

The prototype is 3D-printed with thermoplastic polyurethane (TPU) $95A$ with Bambu X1-Carbon Printer. $20\%$ of the periphery is cut off and replaced by a wrinkled silicone shell to increase friction. This is inspired by skin patterns observed in real elephants\cite{schulz2024elephants}. Wrinkles are designed by randomly distributing horizontal cuts on the mold. The shell is then fabricated via molding of Smooth-On Ecoflex™ 00-30 silicone. 

The tip gripper is also 3D printed. The three fingers are symmetrically distributed at the end of the robot arm at an interval of $120^\circ$. Compared with the two-finger gripper, a three-finger configuration possesses a more stable grasping ability. A rotating mortise is designed on the base of the gripper to connect with the tip of the robot arm.

Serving as the connection between the rigid and soft components, the motor housing contains the actuation mechanisms, which include three GM6020 motors and a DS-929MG steering gear. These actuators regulate the tension of the ultra-high molecular weight polyethylene cables (UHMWPE) that run through the manipulator. Three GM6020s are distributed symmetrically at intervals of $120^\circ$, controlling three individual cables running through the peripheral of the soft manipulator. Winders are installed in the motor driver, and each cable can be wound around each winder to tighten and relax without idling. 

The tip gripper, on the other hand, is actuated by the DS-929MG servo motor, which is connected to two UHMWPE cables that are braided with three strands. The cables run from the central axis of the motor housing through the cavity in the soft arm's central backbone. When approaching the end-effector, the cables split into three strands, and each group of strands goes through the dorsal side and ventral side of the three fingers, respectively. Refer to figure \ref{Fig1}, two groups of cable are respectively tied on both sides of a double-arm-connector that connects to the steering gear. By rotating clockwise and anti-clockwise, one group of cables would be tightened while the other group would be relaxed. With this arrangement, three fingers can be actuated together for opening and closing by two groups of cable.

 

\subsection{Kinematic Model}
\label{kinematic}
In this section, we establish the forward kinematic model in three steps. Firstly, we analyze the gravitational forces acting on the manipulator when the base section of the soft manipulator is at an oblique angle. Secondly, we actively actuate a single cable to induce a bend. Lastly, based on the single cable actuation model, an additional second cable is tightened by a different length to induce a twist.

\subsubsection{\textbf{Resting Under Gravity}}
The soft arm is mounted on a 7-DoF rigid robot. As the end-effector pose of the rigid arm changes, it will form an initial inclination angle relative to the original vertical plane. Therefore, the same cable actuation could induce different shape configurations because of gravity. This necessitates analyzing the resting state of the soft arm under the influence of gravity with respect to the initial inclination angle of the base.

We define the initial static state as the set of joint angles in each section when the soft arm passively hangs downward under gravity without any cable actuation. $\theta_0$ is defined as the angle between the central axis of the soft arm and the horizontal line shown in Fig. \ref{fig4} (b).



According to the equation of elementary beam theory, the relationship between an applied bending moment $M$, Young's Modulus $E$ ($E=16.9MPa$ for our soft manipulator), the second moment of area $I$, and the resulting curvature $\kappa$ of the beam can be described as equation \ref{eq2}:

\begin{equation}\label{eq2}
    M=EI\kappa
\end{equation}

The gravitational force exerted on each joint induces a moment $M$ and a curvature $\kappa$. Taking the case of a two-section-arm as an example, as shown in Fig.\ref{fig4} (b), given the mass and center of gravity (CG) of the first and second section, the gravitational moment of the second section about $joint 1$ (shown in blue rectangle) can be expressed $m_1 g a_1 K_c \cos(\theta_0 + \theta_1)$, where $K_c$ is a constant representing the location of CG on the trapezoid base. Thus, if a certain tilt angle $\theta_0$ is given,  $\theta_1$ can be calculated by:

\begin{equation}
m_1 g a_1 K_c \cos(\theta_0 + \theta_1) = EI \frac{\theta_1}{L_1}
\label{eq: gravity_moment}
\end{equation}

When there are $N$ joints in total (Fig. \ref{fig4}(c), $N=17$ in our case), when considering $joint_n$, we can effectively treat the remaining $N-n$ sections as one entire part, and derive its gravitational moment about $joint_n$. Subtracting and simplifying two adjacent equations as \ref{eq_gravity}, the joint bending angles $\theta_1 \sim \theta_N$ can be solved.

\begin{equation}
\left\{
\begin{aligned}
    & EI \frac{\theta_{n}}{L_{n}} = m_{n} g a_{n} K_c \cos\left(\theta_0 + \sum_{i=1}^{N} \theta_i\right) \\
    & EI \frac{\theta_n}{L_n} - EI \frac{\theta_{n+1}}{L_{n+1}} =\\&\left(m_n K_c + \sum_{i=n+1}^{N} m_i \right) g a_n K_c \cos\left(\theta_0 + \sum_{i=1}^n \theta_i\right)
\end{aligned}
\right.
\label{eq_gravity}
\end{equation}

Each joint has a maximum rotation limit of $\Delta \theta = 30^\circ$. If gravity causes the joint to reach this limit under a certain tilt angle (which generally happens in the first joint), these two adjacent sections would be treated as a rigid body. Therefore, the gravity torque equation shall be updated. Taken $joint 1$ as an example, if $\theta_1 = 30^\circ$, the remaining 16 joints $\theta_2 \sim \theta_{17}
$ can be obtained by the remaining 16 equations. 

Finally, the joint angle $\theta_1 - \theta_{17}
$ can be used to determine the cable length in the resting state, preparing for the subsequent calculation under cable actuation.

\subsubsection{\textbf{Bending}}
\label{sub_bend}
Since each cable runs through the manipulator from the tip to the base, if one cable is shortened by the amount of $\Delta X$, it would affect the cable length in every single section by $\Delta x$. Since every section and joint is $k_p = 0.9196$ times the former one, we can derive the relationship of $\Delta X$ and $\Delta x$. Consequently, the core solution for establishing this single-segment length-based kinematic model is to assign proper cable length to each joint based on the special geometry of the logarithmic spiral when a certain amount of cable actuation is given. 

As Fig.\ref{fig4} (d) shows, when actuated, the cable lengths $AB$ and $CD$ stay constant within the section, while the cable length between $BC$ changes. Thus, the main purpose is to derive how the changing length of a certain cable $D_1$ would affect the cable length $x_i = BC$ between each joint. 

According to equation \ref{eq2}, if we regard each joint as a small cylinder, when the joint deforms, the backbone of the small cylinder can be approximately regarded as an arc with a constant length $L$. Thus, the second moment of area of the joint $I$ and the curvature $\kappa$ can be described as:

\begin{equation}
I=\frac{\pi}{4}r^4\label{eq3}
\end{equation}

\begin{equation}
\kappa=\frac{\Delta \theta}{L}\label{eq4}
\end{equation}

According to the logarithmic-spiral proportionality, $r_i={k_pr}_{i-1}$, $L_i={k_pL}_{i-1}$. We assume the torque relationship between each section empirically as:

\begin{equation}
M_i=k_p^{-\frac{5}{2}}M_{i-1}\label{eq5}
\end{equation}



Substitute equation \ref{eq3}, \ref{eq4}, \ref{eq5} into equation \ref{eq2}, the joint rotation described in $\Delta \theta_{i}$ can be determined. To establish relations between joint angles $\theta$ and cable length $X$, the cosine theorem can be applied as:

\begin{equation}
\theta = \cos^{-1} \left( \frac{R_\text{hole}^2 + R_\text{hole}^2 - X^2}{2 \cdot R_\text{hole} \cdot R_\text{hole}} \right) 
\end{equation}

Where $R_\text{hole} = BE = CE$, $X = BC$. The current joint angle $\theta$ is the difference between the initial angle and the rotation angle $\theta = \theta_\text{initial} - \Delta \theta$, while the current cable length $X$ is the difference between initial length and the changing length: $X = X_\text{initial} - \Delta x$:


The ratio $portion_i$ can be calculated to distribute the input total cable length $L$ and distribute $L_i$ to each joint as:

\begin{equation}\label{eq17}
portion_i = \frac{\Delta X_i}{\sum \Delta X_i}, \quad L_i = L \cdot |portion_i|
\end{equation}

Utilizing the calculated $portion_i$, Algorithm.\ref{algo1} computes the final joint pose given the desired length change for a particular cable:

\begin{algorithm}
\caption{Calculate length distribution for bending}\label{algo1}
\textbf{Input:} $D$ = the length change for a $cable$ \\
\textbf{Output:} length distribution\\

\textbf{Definitions:}
\begin{itemize}
    \item \textbf{n}: The total section number of the manipulator.
    \item \textbf{limitation L'}: The cable length between each section when the joint is not bending.
\end{itemize}

\begin{algorithmic}[1]
\State $portion($i$) = \frac{\Delta X_i}{\sum \Delta X_i}$
\For{$i = n : 1$}\label{algln1}
        \State $distribute Length($i$) = portion ($i$) * D$
        \If{$distribute Length($i$) \le limitation L'($i$)$}
        \State $newLength($i$) = limitation L'($i$) -distributeLength($i$)$
        \Else
        \State $newLength($i$) = 0$
        \State assign the additional length to the rest of the joint
        \EndIf
\EndFor
\State calculate joint pose via $newLength$
\end{algorithmic}
\end{algorithm}

Since the bending movement would only be performed in planes, the shape of the trunk can be determined by the rotation angles $\Delta \theta_{i}$ of the joints via transformation matrix described in equation \ref{eq7} below, where the $x$-direction is set along the trunk body:

\begin{equation}
T_{0}^{i}=\prod_{1}^{i}\left[\begin{array}{cccc}
\cos \left(\Delta \theta_{i}\right) & -\sin \left(\Delta \theta_{i}\right) & 0 & {x_1\bullet k}_p^{i-1} \\
\sin \left(\Delta \theta_{i}\right) & \cos \left(\Delta \theta_{i}\right) & 0 & 0 \\
0 & 0 & 1 & 0 \\
0 & 0 & 0 & 1
\end{array}\right]\label{eq7}
\end{equation}

\subsubsection{\textbf{Twisting}}
A twist is induced by actively tightening the second cable by a different length on top of the actuation length of the first cable. 

When the second cable is actuated, the rotation axis of the affected sections changes. The gradual change in the rotation axis causes the manipulator to rotate progressively out of the initial plane, resulting in a twist effect as shown in Fig.\ref{fig4} (e). 

We model the influence of the two cables in a sequential manner. Firstly, tightening the first cable by a length $D1$ induces a bend in a specific plane. At this stage, some distal joints may reach their maximum allowable angles, while others remain within their limits. Subsequently, when the second cable is tightened by an additional length $D_2$, joints that have already reached their limits remain unaffected due to the constraint imposed by the first cable. The second actuation thus affects only those joints with remaining rotational capacity. Each section rotates about its specific axis, following the same distribution principle described in Equation \ref{eq17}. 

Consequently, we can approximately regard the groups of maximally rotated sections as rigid bodies and calculate the rotation of the rest of the sections. The detailed process is demonstrated in Algorithm \ref{algo2}.

\begin{algorithm}
\caption{Calculate joint pose for twisting}\label{algo2}
\textbf{Input:} $D1$ = the length change for $tendon1$, $D2$ = the length change for $tendon2$\\
\textbf{Output:} Joint pose\\

\textbf{Definitions:}
\begin{itemize}
    \item \textbf{n}: The total section number of the manipulator
\end{itemize}

\begin{algorithmic}[1]
\State calculate $originalL2$ after $D1$ is pulled via $\textbf{Algorithm 1}$
\State find the joint indice $m$ where all joints after reaching maximum bending angles 
\State $portion($i$) = \frac{x_i}{\sum_{i=1}^n x_i}$
\For{$i = n : 1$}\label{algln2}
    \If{$i > m$}
        \State $newL2($i$) = originalL2($i$)$
    \Else
        \State find the rotation axis for each single joint
        \State find the maximum rotation angle $\theta max($i$)$ when two links collide
        \State find the maximum length change $D2 max($i$)$ via $\theta max($i$)$
        \State $distribute L2($i$) = portion ($i$) * D2$
        \If{$distribute L2($i$) \le D2 max($i$)$}
        \State $newL2($i$) = originalL2($i$) - distributeL2($i$)$
        \Else
        \State $newL2($i$) = D2 max($i$)$
        \State assign the additional length to the rest of the joint
        \EndIf
    \EndIf
\EndFor
\State calculate new joint pose via $newL2($i$)$
\end{algorithmic}
\end{algorithm}


\subsection{Model Validation via Motion Capture System}\label{method_mocap}


\subsubsection{Bending}
For each trial, one cable running through a vertex of the trunk was incrementally tightened while the other two cables were relaxed. The tightening process was conducted in 26 steps, with each step shortening the length of the active cable by $5 mm$. The lengths by which the other two cables were relaxed at each step were computed using Algorithm \ref{algo1}, ensuring the manipulator reproduced the target shape as closely as possible.


\subsubsection{Resting under gravity}
To validate the ability of our soft manipulator to replicate the bending behavior of an elephant trunk at rest under gravity, we measure its deformation at various base angles. The manipulator was securely mounted on an adjustable-angle wedge mounting plate, allowing precise control of the base rotation angle. The plate was configured to tilt the manipulator to angles of $0^\circ$, $45^\circ$, $90^\circ$, and $135^\circ$ relative to the vertical axis. In each configuration, the cables running through the vertices of the manipulator were left relaxed, enabling the manipulator to bend freely under its weight until a static equilibrium shape was reached.


The captured data included the positions of the 15 rigid body frames, allowing precise reconstruction of the manipulator's backbone shape. All data were saved in CSV format for post-processing and analysis. The recorded bending profiles were then compared to theoretical predictions from the physics model described in Section \ref{kinematic}.

\subsection{Measuring Sources of Error}
\label{subsec: SoE}
There are two main sources of error that could affect the accuracy of the data captured using the motion capture system.
\subsubsection{Material Elasticity}
A systematic source of error in our experiments arose from the elasticity of the material used to 3D-print the soft manipulator, which violated the assumption that each link behaves as a rigid body without deformation. This elasticity introduced residual slack in the cables during actuation. To account for this error, we fully wrapped the manipulator and measured the residual slack in each cable. These measurements were then used to compute compensation factors for the cable lengths, enabling adjustments to the tightening and relaxing steps. By compensating for material deformation, we improved the accuracy of the manipulator’s motion and alignment with the theoretical forward kinematic model.

The total measured slack in the cable ($L_{slack}$) was determined by fully wrapping the manipulator and recording the excess cable length required to achieve the desired shape. This slack was then evenly distributed across the total number of tightening steps ($N_{steps}$), resulting in a stepwise compensation factor for each cable:

\begin{equation}
    C_i = \frac{L_{slack}}{N_{steps}}
\end{equation}

To ensure progressive correction, we incorporated the compensation factor into each step of cable tightening, such that the adjusted tightening length at step $i$ was computed as

\begin{equation}
    \Delta L'_i = \Delta L_i + C_i \cdot i
\end{equation}

where $\Delta L_i$ is the theoretical length based on our analytical model. This linear scaling of the compensation factor with the step index ensures that the cumulative slack is systematically offset as the manipulator bends further.


\subsubsection{Material Deformation and 3D Printing Quality}
The 3D-printed manipulator, fabricated using TPU, exhibited slight asymmetries due to the material's high liquidity during the printing process. TPU tends to spread before solidifying, and under the influence of gravity, the printed structure may develop minor deformations. These asymmetries caused the manipulator to deflect slightly to one side during actuation. To account for the random error that could be introduced into the results due to this defect, we conducted our bending experiment 2 times and took the average of each rigid body marker position at each step.


\section{Data Availability}
All data needed to evaluate the conclusions in the paper are present in the paper or the Supplementary Materials. The data generated in this study are provided in the Source data file.

\bibliography{bib}

\section{Acknowledgements}
The authors are grateful to Dr. Li Yangfan and Mr. Lu Dingjie for their insightful discussions and helpful suggestions that greatly contributed to the development of this paper. This research is supported by the National Research Foundation (NRF), Prime Minister’s Office, Singapore under its Campus for Research Excellence and Technological Enterprise (CREATE) programme. The Mens, Manus, and Machina (M3S) is an interdisciplinary research group (IRG) of the Singapore MIT Alliance for Research and Technology (SMART) centre. This research is also supported by A*STAR, Singapore, through the Italy-Singapore collaborative project ``DESTRO - Dextrous, strong yet soft robots" (C.L.).

\section{Author Contributions}
H.H. and H.W. conceived and designed the study. C.F. investigated the optimal parameters in the logarithmic spiral. H.H. and C.F. established the forward kinematic model and designed the hardware. M.Y. and R.X. developed the embedded control system for the soft manipulator. H.W. integrated the control of the soft and rigid components into one system. H.H., H.W., M.Y., and R.X. completed the motion capture validation experiment. H.H., H.W., C.F., M.Y. reproduced the demonstration of elephant grasping strategies. Y.Z. provided insights on hardware prototyping and gave constructive advice for this work. Z.W. provided design and fabrication advice for the soft manipulator prototype. H.H., W.H., C.F., and F.Y. analyzed the data and drafted the manuscript. J.L., C.L., AND M.A. supervised and revised the research. All authors reviewed the manuscript.
\end{document}